\RequirePackage{fix-cm}

\documentclass{article}  

\usepackage{graphicx}
\usepackage{amsmath}
\usepackage{bm}
\usepackage{amsfonts}

\begin{document}

\title{The Embeddings World and Artificial General Intelligence}



\author{Mostafa Haghir Chehreghani \\
              Department of Computer Engineering \\
              Amirkabir University of Technology (Tehran Polytechnic), Tehran, Iran \\
               mostafa.chehreghani@aut.ac.ir          
}



\date{}

\maketitle

\begin{abstract}
From early days, a key and controversial question inside the artificial intelligence community was
whether Artificial General Intelligence (AGI) is achievable.
AGI is the ability of machines and computer programs to achieve human-level intelligence and do all tasks that a human being can.
While there exist a number of systems in the literature claiming they realize AGI,
several other researchers argue that it is impossible to achieve it.

In this paper,
we take a different view to the problem.
First, we discuss that in order to realize AGI, along with building intelligent machines and programs,
an intelligent world should also be constructed which is on the one hand, an accurate approximation of our world and on the other hand,
a significant part of reasoning of intelligent machines is already embedded in this world.
Then we discuss that AGI is not a product or algorithm, rather it is a continuous process which will become more and more mature over time
(like human civilization and wisdom).
Then, we argue that pre-trained embeddings play a key role in building this intelligent world and as a result, realizing AGI.
We discuss how pre-trained embeddings facilitate achieving several characteristics of human-level intelligence,
such as embodiment, common sense knowledge, unconscious knowledge and continuality of
learning,
 by machines.

\paragraph{Keywords.}Philosophy of artificial intelligence; artificial general intelligence (AGI);
embedding (representation);
pre-trained embeddings;
intelligent world;
human intelligence;
machine intelligence.

\end{abstract}

\section{Introduction}
\label{intro}

In recent years, advanced machine learning (ML) and
artificial intelligence (AI) techniques have become very popular,
due to their high performance in solving problems from different domains.
These techniques are successful in solving one specific problem (or a few problems from one specific domain),
such as chess-playing, car driving and mathematical calculations.
However, for example a computer program that can play chess at a very professional level,
might not be capable of playing another game,
even at a beginner level.
For each individual and independent task, usually a separate algorithm is needed to be developed.

When the field of artificial intelligence first emerged,
the original focus of researchers was, however,
Artificial General Intelligence (AGI).
AGI is the ability of machines and computer programs to solve all the tasks that a human being can.
Searle~\cite{searle_minds_1980} is one of the first researchers who makes a distinction between:
AGI which implies that the intelligent machine does not simulate a human mind,
but actually has a mind and is capable of an intelligence equal or superior to human beings;
and narrow AI which implies that a computer program is capable of performing only one specific task or a few relevant tasks\footnote{In the literature,
researchers use terms such as {\em strong AI} and {\em general AI} to respectively refer to human-level intelligence and the capacity of solving many tasks,
opposed to {\em weak AI} and {\em narrow AI} that respectively refer to an intelligence
that do not experience human consciousness and an intelligence that is applicable to only a very specific setting.
In this paper, by AGI we mean an intelligence which is on the one hand capable of solving several tasks from different domains and on the other hand,
at least to some degree achieves key characteristics of human intelligence.}.


%

From early days, a controversial question
inside the AI community was whether AGI is achievable by machines and computer programs.
There are several attempts in the literature to answer this question, which can be divided into optimistic and pessimistic viewpoints.
A group of researchers, mostly from AI community, believe that achieving AGI is feasible,
and try to develop systems or algorithms that realize AGI~\cite{Hutter:04uaibook,DBLP:journals/jsait/CohenVH21,sci,GPT-3,DBLP:conf/nips/VaswaniSPUJGKP17}.
While most of well-known algorithms are based on reinforcement learning~\cite{Hutter:04uaibook,DBLP:journals/jsait/CohenVH21},
a few others use probabilistic approaches~\cite{sci} or deep neural models~\cite{GPT-3,DBLP:conf/nips/VaswaniSPUJGKP17}.
However, it is a common belief that we are still far from the point of having systems that  realize AGI and work in practice.
Even known systems such as XIAI~\cite{Hutter:04uaibook} or more recent and improved approaches~\cite{DBLP:journals/jsait/CohenVH21} are not practical.
Some other researchers that mostly have a philosophical view to the problem, criticise AGI and
believe that due to characteristics that human intelligence has (embodiment, common sense knowledge, tacit and unconscious knowledge etc),
it is not possible to develop human-level intelligence for machines and computer programs~\cite{searle_minds_1980,Alchemy,Dreyfus92,10.5555/7916,10.5555/540148,odifreddi_1997,Mantaras,Fjelland}.

However, we believe before trying to answer this question,
we should first elaborate on its different aspects.
First, AGI should not be seen as an algorithm or a product or a computer program,
rather, as a continuous process within agents and systems become more and more intelligent over time,
as human intelligence is a continuous process.
Second, human intelligence is affected by human's mental model of the world and as a result, by our  world.
For example, a characteristic of human intelligence is {\em unconscious knowledge} that consists of facts such as
"water always flows downward"~\cite{Mantaras}.
However, such facts are derived from our world and our experiences in this world.
So, for example in a world without gravity, this statement would not be a part of human's unconscious intelligence,
but rather some meaningless statement.
In a similar way,
in order to realize AGI, we first need to build an {\em intelligent world}
which is on the one hand, an accurate approximation of
our world and on the other hand, a significant part of knowledge and reasoning of intelligent
machines is already embedded unconsciously in this world.
Machines should be able to obtain this unconscious knowledge from this world quickly and with almost no effort.
Third, the answer to the above question should not be simply yes or no.
It is more reasonable to provide a list of characteristics that human intelligence has,
and discuss what characteristics are achievable by machine intelligence and to what extend.

In this paper, we study achievability of AGI,
while taking into account these considerations.
First, we introduce the notion of {\em embeddings world}
consisting of pre-trained embeddings of objects,
as a trained approximation or abstraction of our world which is proper for machines and computer programs.
Embeddings are vectors in a low-dimensional vector space,
that represent objects of a domain so that  their structural relations are preserved.
Pre-trained embeddings are embeddings that are learned and computed usually using a general-purpose objective function.
Then, they are stored and can be used, as good quality representations of real objects,
to solve several tasks.
Second, we argue that along with other techniques such as reinforcement learning that are discussed in the literature,
pre-trained embeddings provide a path toward AGI.
More precisely, we discuss how pre-trained embeddings facilitate achieving several characteristics of human intelligence,
such as embodiment, common sense knowledge, unconscious knowledge, and continuality of learning,
by machines.

The rest of this paper is organized as follows.
In Section~\ref{sec:relatedwork},
we provide an overview on related work.
In Section~\ref{sec:embedding}, we define the embeddings world and discuss why it can be useful in realizing AGI.
In Section~\ref{sec:agi}, we discuss that the embeddings world can help machines to achieve several characteristics of human-level intelligence.
Finally, the paper is concluded in Section~\ref{sec:conclusion}.

\section{Literature review}
\label{sec:relatedwork}

In the literature, on one hand there exist a number of algorithms and systems that claim that they realize AGI.
On the other hand, there are philosophical discussions that argue that achieving AGI is impossible, or very hard.
In this section, we briefly review these two types of related work.

AIXI~\cite{Hutter:04uaibook} is one of famous algorithms for AGI, which is based on reinforcement learning.
It maximizes the expected total rewards received from the environment.
In each time step, it checks every possible hypothesis and evaluates how much reward that hypothesis would generate depending on the next action taken.
This reward is weighted by a belief
which is computed from the length of the hypothesis: a longer hypothesis is considered more complicated than a shorter one and as a result, less likely.
AIXI selects the action that has the highest expected weighted reward.
Since at each step AIXI considers and evaluates all possible hypothesizes,
it is not computationally practical.
Moreover, the agent may hijack its reward.

The literature includes several improvements of AIXI.
One of the most recent algorithms is the Boxed Myopic Artificial Intelligence (BoMAI) method~\cite{DBLP:journals/jsait/CohenVH21},
which is also based on reinforcement learning and tries to maximize reward, {\em episodically}:
it is run on a computer which is placed in a sealed room with an operator.
When the operator leaves the room, the episode ends.
The intelligent agent can perform any task, except hijacking its reward.
The authors believe that this procedure provides a path to AGI, despite the dangerous failure of reward hijacking.
However, similar to AIXI, BoMAI is not practical.

One of recent successes in producing human-quality output by a computer program is the
Generative Pre-trained Transformer 3 (GPT-3) program~\cite{GPT-3}.
GPT-3 is a language model developed by OpenAI, that uses deep learning to produce human-style text.
The quality of the text generated by GPT-3 is high so that it is difficult to distinguish it from human-written text.
The Gato system~\cite{DBLP:conf/nips/VaswaniSPUJGKP17} introduced by DeepMind,
is a general-purpose system that can perform 604 tasks of different types,
including captioning images, chatting and conversation, stacking blocks with a real robot arm and playing Atari games.
It uses transformers for learning and inference, which are
deep learning models that use the attention mechanism for weighting the significance of each part of the input data, differently~\cite{DBLP:conf/nips/VaswaniSPUJGKP17}.


Searle is one the first scientists who criticizes AGI (and AI in general),
and believes that achieving AGI is impossible~\cite{searle_minds_1980}.
Another famous critic of AGI is Hubert Dreyfus, who criticises the philosophical foundations of AGI.
In his several books and writings, including Alchemy and AI~\cite{Alchemy}, What Computers Can't Do~\cite{Dreyfus92} and Mind over Machine~\cite{10.5555/7916},
he argues that human intelligence depends mainly on unconscious knowledge and decision making,
rather than conscious knowledge.
He believes that computers and AI programs can never fully capture this unconscious or background knowledge,
and they are unable to perform fast decision making and problem solving,
which is based on unconscious knowledge.
Therefore, computers can never achieve human-level intelligence.

Weizenbaum~\cite{10.5555/540148} states that
computer intelligence and human reasoning are fundamentally different.
Computer power is the ability to perform computations at a very high speed,
whereas human reasoning consists of prudence and wisdom.
Prudence is the ability of making proper decisions in
concrete situations, and wisdom is the ability of seeing the whole picture.
These capabilities are not computational or algorithmic, so,
can not be simulated by a computer.
In a similar way, Penrose~\cite{odifreddi_1997} argues that human thinking is essentially different from computer programs.

Mantaras~\cite{Mantaras} looks at recent advances in AI, that are based on the analysis of large volumes of data and big data processing.
He discusses challenges and difficulties of this approach and highlights that
in order to realize GAI, there is still the need to provide common sense knowledge to computer programs.
In another recent paper, Fjelland~\cite{Fjelland} reviews important critiques on AGI and
argues that despite considerable developments of AI for specific tasks, we
have not come much closer to achieving AGI.
The author revives Dreyfus' argument that computers are not in the world and therefore,
achieving AGI is in principle impossible.


\section{The embeddings world}
\label{sec:embedding}

State of the art recent AI and ML techniques for solving different problems  rely on learning or computing an embedding (representation)
for each data-point or object.
An embedding is a vector in a low-dimensional vector space that encodes the whole information of an object and its relations to other objects in the domain,
as much as possible\footnote{Note that
there are other notions of {\em embedding} in AI, e.g., the notion of {\em embedding} in {\em graph pattern mining} \cite{DBLP:journals/fgcs/ChehreghaniABB20,DBLP:journals/datamine/ChehreghaniB16,DBLP:journals/tsmc/ChehreghaniCLR11},
that should not be confused with the notion of {\em embedding} used in this paper.}.
However, what makes it more effective than a typical encoding is that it is {\em learned} to preserve similarities, properties and structural relationships
among objects.
For example, in embedding learning for words, two words that have similar syntaxes or semantics, find similar embeddings, i.e.,
are mapped to closer point in the vector space  than two words that are irrelevant.
In a similar way, phrases, sentences and documents also find such relation preserving embeddings.
As an another example, consider a social network wherein each person is modeled by a node in a graph and different relationships among people are modeled by graph edges of various types. Nodes (individuals) that have similar properties (profiles, etc) and similar structural positions in the graph (have same friends, belong to same communities, etc),
find similar embeddings.


Embeddings have several advantages that make them useful for AGI.
\begin{itemize}
\item
First, {\em pre-training} them makes them computationally easy and accessible to use.
Pre-training means instead of learning embeddings for a specific task, they are trained using a general and task-independent objective function.
This objective function could be defined, for example, using an encoder and a decoder.
The encoder maps an object/data-point to its embedding and
the decoder retrieves it from its embedding.
The objective function, that should be minimized, might be defined as the sum of the distances between the original objects and their retrieved versions by the decoder.
The generated embeddings are stored and used to represent objects in different learning tasks.
Generating and using task-specific embeddings, wherein the embeddings are learned  using a specific objective function derived from the learning task,
usually yields a slightly better result for that specific task.
However, general-purpose pre-trained embeddings address several issues.
On the one hand, learning embeddings can be very time and resource consuming.
Using pre-trained embeddings significantly improves the efficiency of the learning process.
On the other hand, in supervised tasks such as classification,
a huge labeled dataset is required to have a good performance.
This labeled data is used to define the task-specific objective function.
In pre-training, there is usually no need to labeled data.
\item
Second, they provide a good abstraction or approximation of objects in a specific domain and their relations.
High quality embeddings are generated in different domains, such as (but not limited to):
\begin{itemize}
\item
words and other textual objects such as phrases, sentences and documents using e.g., Bert~\cite{textemb,bert},
\item
graph objects such as social networks, technological networks, collaboration networks and their components (nodes, edges, subgraph)~\cite{gcn,nmi},
\item
images~\cite{imagenet}, videos~\cite{videoemb} and speech~\cite{speechemb}.
\end{itemize}
Different forms of neural models are usually used to learn or generate embeddings.
They generate multi levels of embeddings, so that
each layer
transforms the embedding at one level
into an embedding at a higher and slightly more abstract level~\cite{lecun2015deep}.


We believe extending domain-specific embeddings to a general domain will be feasible in future.
Suppose as an example, we have a picture and a paper on a table.
The projection of this situation in the embeddings space consists of three embeddings of the objects picture, paper and table
so that each embedding comes from its own specific domain.
Furthermore, there should be embeddings that reflect the information of relative positionings of the objects against each other.
We may call these embeddings as {\em connectors} and in this example, we will have three of them
that depict the relative positioning of paper and table, the relative positioning of picture and table and the relative positioning of paper and picture.
While {\em connectors} can be simply some codes to distinguish different situations, in a better way they can be learned so that
similar connectors find similar embeddings.
In this way, embeddings can finally provide an almost {\em exhaustive} and {\em accurate} model of our world.


\item
The third advantage of embeddings is that they can be computed {\em inductively}.
This means we do not need to use the whole of data to learn/compute embeddings.
Also, when some new data arrives or our existing data changes, we do not always need to redo our embedding learning process,
to find embeddings for new data.
What we need is that we assume the embedding of an object/data-point
is a {\em parameterized} function of its features.
This function can be defined using e.g., a neural network.
Then the task is to learn the parameters, i.e., find proper/optimal values for them.
This can be done using only a part of data.
Then, after learning parameters, for each object/data-point, using its features as the input of the function and the learned values of the parameters,
its embedding is computed very quickly.
In this case, instead of storing embeddings of all objects/data-points,
it is sufficient to store the learned values of parameters.
\end{itemize}

With these
{\em pre-trained}, {\em general-purpose}, {\em exhaustive}, {\em accurate} and {\em inductive}
embeddings in hand, simple algorithms are usually sufficient to obtain good results for different learning tasks.
In fact, in this type of AI, the main part of intelligence is laid in the embedding and representation generation phase,
to create an accurate approximation of the world for computer programs.
After creation of this world, which we call the {\em embeddings world},
computer programs and algorithms can work efficiently and accurately in it,
and can be easily extended to do several different tasks.

\section{Realizing human intelligence}
\label{sec:agi}

Researcher pessimistic about AGI believe that human intelligence has characteristics such as
embodiment, common sense knowledge and unconscious knowledge,
that are dedicated to humans; and machine intelligence can not achieve them.
In this section, we discuss how pre-trained embeddings may help an intelligent machine to achieve (at least to some degree)
these characteristics.

\subsection{Embodiment}
One of the strongest critiques of AGI is that computers and intelligent programs are not embodied and are not in the world~\cite{Dreyfus92}.
So they work on an abstraction of the world, codified by a language that represent the surrounding information,
whereas humans have direct experiences of their surroundings, and actually are {\em situated} in the world.
Without a body, those abstract representations have no semantic content for computer programs or machines~\cite{Mantaras}.

However, we believe there is an important difference between embeddings and
other encoding methods:
embeddings are {\em learned} ({\em optimized}) to preserve different properties of objects as well as structural and semantic relations among them.
This is not the case for other encodings or representations.
As mentioned in Section~\ref{sec:embedding},
pre-trained embeddings create an accurate and intelligent
approximation of the world for computer programs wherein they can properly act.
So, each object finds a body in this world defined by its embedding,
and relations among objects are also embedded.
An intelligent machine or agent is situated in this {\em embeddings world}:
it itself consists of a set of embeddings and each object surrounding it is also an embedding and connector embeddings can be used to identify
relative positioning, both explicit and implicit, of objects and agents.

\subsection{Common sense knowledge}

Common sense knowledge consists of facts that are known for everyone.
It is thought as a result of our lived experiences.
Examples include: "water always flows downward", "knife cuts cheese, but cheese does not cut knife", and "a pigeon flies, but an elephant does not fly".
Humans can easily and quickly learn and process millions of such common sense knowledge,
and use them for decision making and showing proper reactions.
Equipping machines and computer programs with common sense knowledge is considered as one of the key challenges in AGI~\cite{Mantaras}.

Pre-trained embeddings, along with other tools such as knowledge graphs,
can be used to provide common sense knowledge for intelligent machines~\cite{DBLP:conf/lrec/GoodwinH16}.
A common sense is usually some knowledge about properties of one or more objects or their relationships.
Since embeddings are carefully and smartly computed so that objects' properties are preserved and relevant objects find similar (close) embeddings,
they can carry common sense knowledge.
For example, while the embeddings of words "water" and "downward" are close,
the embeddings of words "water" and "upward" are far.
Therefore, "water" and "downward" may form a common sense knowledge.
We may state that in its simple form, in the embeddings world
a common sense knowledge consists of a set embeddings that are very close to each other.

There are a few key points, here.
First, {\em pre-training} of embeddings makes it feasible to {\em quickly} find close embeddings that form a common sense knowledge.
To obtain common sense knowledge, an intelligent machine may only need to perform operations such as computing the distance between two embeddings
or finding embeddings within a given radius.
Such operations can be done much faster than operations such as learning optimal embeddings,
not to mention they can be done within a pre-processing phase too.
In the case of pre-trained {\em parameterized} embeddings, embeddings of objects can be computed very quickly using the learned parameters.
So in this case, common sense knowledge will be available very fast too.
This is consistent with an interpretation of common sense knowledge of human beings,
that humans obtain this knowledge instantaneously so that it is felt that they already have it.

Second, common sense knowledge of humans has characteristics such as:
i) it is not a symmetric relation, e.g.,
the common sense knowledge that adult people have about water (or even about babies) is not the same as the common sense knowledge that water or babies have about adults,
ii) it is not necessarily limited to humans, for example, some animals may also have forms of common sense knowledge such as
"water flows downward", and
iii) humans (and in general objects of our world) may have different levels of common sense knowledge about other entities,
for example experienced or mature people may have a higher level of this knowledge.
All these characteristics are consistent with the common sense knowledge obtained from pre-trained embeddings.
In the embeddings world, for each object or agent an access level can be defined which determines embeddings accessible to it.
It can consist of its own embeddings and those embeddings that determine its {\em free-will} domain or {\em authority} domain.
Moreover, the object/agent may actually load, process and use only a fraction of these accessible embeddings and over time,
this fraction may become larger and larger.
So, common sense is not symmetric as the embedding of object $O_1$ might be in the authority domain of object $O_2$,
but the embedding of $O_1$ might not belong to the authority of $O_1$.
As soon as the intersection of the authorities of a group of objects becomes non-empty and some of the embeddings in this intersection are close enough,
the objects find common sense knowledge.
The level of common sense knowledge of an object in the embeddings world may depend on its authority and the
fraction of its authority which is loaded and processed.
Also, it may vary over time.

\subsection{Unconscious knowledge}

A characteristics of human intelligence is that in many situations,
they take actions without reasoning and simply choose the appropriate response, without investigating all alternatives and possibilities.
Philosophers such as Dreyfus argue that in these cases humans' intuitions are trained to the point that they can simply "size up the situation"
and show the appropriate reaction.
According to Dreyfus, the human's sense of the situation is based on unconscious and background knowledge about the world.
They use this knowledge to discriminate between what they consider essential and lots of other things that they are aware of, but they consider inessential.
Dreyfus believes that symbolic AI would have difficulties in capturing this unconscious knowledge and
doing the kind of fast problem solving that it allows~\cite{10.5555/7916,Dreyfus92}.

We believe  pre-trained embeddings can help machines and computer programs to capture unconscious knowledge.
Those pre-trained embeddings inside the authority of an  agent that are frequently accessed/processed by the agent or are very close to it,
can form its background and unconscious knowledge.
These embeddings can be stored on a high speed storage device, to be retrieved very quickly by the agent.
Moreover, the results of frequent learning tasks can  be stored so that the agent does not require to run their algorithm each time.
Note that in general, generating high quality embeddings is the most significant and time consuming step  during the learning and reasoning process.
Having high quality embeddings in hand,
a simple learning algorithm that can be run very fast is sufficient to produce accurate results~\cite{10.1109/TPAMI.2013.50}.
In other words, the whole learning process is so slow and time consuming, so that in order to have unconscious knowledge and fast problem solving,
using pre-trained embeddings seems to be inevitable.
In fact, pre-trained embeddings already carry a considerable amount of learned knowledge and give it to agents and computer programs.

\subsection{Learning and reasoning}

\begin{figure}
\centering
\includegraphics[scale=0.6]{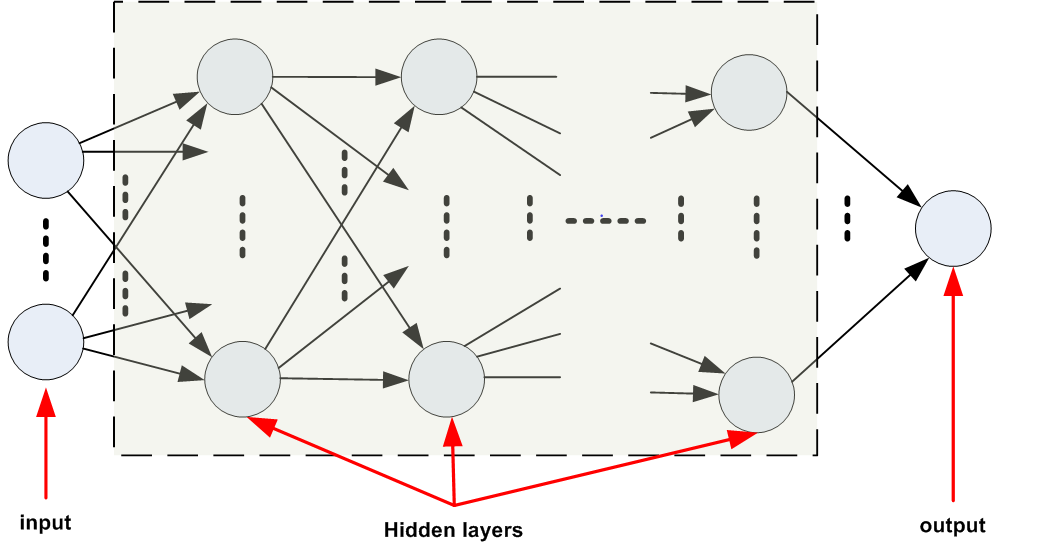}
\caption{\label{fig:emb}In this simple neural network, each hidden layer produces an embedding of the input object.
If we produce most of hidden layers during pre-training (using a general-purpose objective function),
one or a few layers will be left for the learning and reasoning phase, which can be done very efficiently.}
\end{figure}

State of the art algorithms for learning and reasoning in different domains,
including speech recognition~\cite{speechemb,speechemb2} natural language processing~\cite{bert,nlp} and graph data analysis~\cite{gcn,nmi,gatv2},
are based on computing embeddings for input objects.
Instead of computing these embeddings when solving an specific task,
we can pre-train them using a general-purpose objective function.
Then, for each task a simple method that instead of objects, takes their pre-trained embeddings as input,
usually works fine~\cite{10.1109/TPAMI.2013.50}.
Figure~\ref{fig:emb} shows an example of a neural network that consists of the input layer, the output layer and a number of hidden layers.
Each layer takes the output of its previous layer as input and using some trainable parameters, generates an embedding.
The hidden layers (or at least some of them) can be pre-trained, so that only the output layer is trained during an specific learning task.
This may slightly decrease the quality of the results, but enormously improves the efficiency and run time of the learning process.

A domain that particularly depicts the strength of pre-trained embeddings, is {\em neuro-symbolic AI}.
Neuro-symbolic AI states that combining deep neural models with classical rules-based symbolic AI can improve the learning and reasoning process.
More precisely, it suggests to use deep learning for feature extraction and embedding learning,
and rule-based symbolic AI to manipulate and reason with these embeddings~\cite{DBLP:journals/corr/abs-2109-06133}.
Recent studies~\cite{Mao2019NeuroSymbolic} show that for tasks such as visual reasoning~\cite{DBLP:conf/cvpr/JohnsonHMFZG17},
neuro-symbolic models outperform deep models, when a limited training dataset is available.
We think one of  key reasons for the success of neuro-symbolic AI is the high quality embeddings that are learned to
capture and encode different aspects of objects, and make it easy to reason about them.


\subsection{Modeling infinity}

One of challenges that AGI encounters, is modeling the environment of an agent.
The environment of an agent which is the whole or part of our world,
can consist of millions of millions and even an infinity many number of objects.
This renders the algorithms inefficient and also, makes storing information required for modeling such an environment impossible.
So, it is required for computer programs  to find an efficient feasible solution to deal with this infinite number of objects.

In pre-training, embeddings are usually generated and stored, so that an intelligent agent just needs to use them.
In some cases, however, the number of such embeddings can go to infinity as the number of objects is infinite.
For example,  the number of words in a language (English for instance) is finite.
Even sometimes, a dictionary is used that consists only of words in the language that are important.
Then, embeddings are computed and stored for all words or for words in the dictionary which nevertheless, form a finite set.
However, the set of all images or the set of all graphs are very large or even infinite.
So in such cases, it is impossible to store the pre-trained embeddings of all images or all graphs.
A technique which is useful in such cases is {\em inductivity}.
As discussed in Section~\ref{sec:embedding},
in inductive machine learning, part (not the whole) of the training dataset is used for training and finding optimal values for parameters.
Then, the learned parameters are used for the whole data.
In computing embeddings, each embedding is defined as a parameterized function of some input features.
During inductive parameter learning, the optimal values of these parameters are learned over some part of the data.
Then these values are used to compute the embedding of any given object from the domain.
As a result, in this case,
instead of storing the embeddings themselves, the much smaller finite set of parameters that are used to compute embeddings, are stored.

\subsection{Continuality of learning}

In most of advanced AI systems that  use deep neural networks, the knowledge and intelligence of machines and computer programs
are grounded in the training data, which is static.
As a result, they are not able to respond correctly to questions and tasks that are time-dependent.
For example, they might answer wrongly to questions like:
who is the current UN secretary-general?
Chen and Liu~\cite{chen2016lifelong} consider {\em lifelong} or {\em continuous} learning as one of the hallmarks of human intelligence,
and mention that current intelligent systems fail to realize it as their learning is mostly {\em isolated}.
They define lifelong learning as a paradigm wherein the intelligent agent learns continuously,
accumulates the knowledge learned in the past and uses it for future learning and problem solving~\cite{chen2016lifelong}.

In the literature of AI research, AGI is mostly thought as a technique or an algorithm or a product, which may or may not be realized someday.
Rather,
we believe AGI is a process, like human civilization, which will be formed continuously and during a long period of time,
and will become more and more mature over time.
In a similar sense, pre-trained embeddings will be computed continuously and will become more and more mature over time.
Changes in the world are reflected in the embeddings world by computing new and updated pre-trained embeddings,
so that the embeddings world will always reflect an updated and accurate approximation of the world.
Moreover, pre-trained embeddings can facilitate continual or lifelong learning.
As stated in \cite{DBLP:conf/aaaiss/SilverYL13}, one of key elements of lifelong learning is
the retention or consolidation of learned task knowledge.
Since pre-trained embeddings are learned and stored once and used repeatedly,
they automatically maintain the learned knowledge.


%

\section{Conclusion}
\label{sec:conclusion}

%

In this paper, we first stated that in order to realize AGI, along with building intelligent machines and computer programs,
an intelligent world should also be constructed which is on the one hand, an accurate approximation of our world and on the other hand,
a significant part of reasoning and intelligence of intelligent machines is already embedded in this world.
Then we discussed that AGI is not a product or algorithm or system, rather it is a continuous process which will become more and more mature over time
(like human civilization and wisdom).
Then, we argued that pre-trained embeddings play a key role in building this intelligent world and as a result, realizing AGI.
We discussed how pre-trained embeddings facilitate achieving key characteristics of human-level intelligence by machines.

Currently, by computing pre-trained embeddings for objects of certain domains
such as texts, images, videos and graphs,  the process of building the embeddings world is already started.
However, we are just in the beginning of this process.
We believe this {\em endless} process will continue and more and more pre-trained embeddings will be generated in more and more domains,
to model our world for intelligent agents as much as possible.
There will be of course several challenges in this process, especially in computational and storage aspects,
and it seems too early to talk about technical details of the actual creation of the embeddings world.
Nonetheless, the way we see the future of AI is that we will have a huge amount of pre-trained embeddings that are continuously completed and updated over time.
Different companies will introduce their own version of the embeddings world, and
most of intelligent computer programs will heavily rely on these embeddings worlds.


\bibliographystyle{plain}
\bibliography{allpapers}

\end{document}